\pdfoutput=1
\documentclass[11pt]{article}
\usepackage{emnlp2021}
\usepackage{times}
\usepackage{latexsym}
\usepackage[T1]{fontenc}
\usepackage[utf8]{inputenc}
\usepackage{microtype}
\usepackage{booktabs}
\usepackage{tabularx}
\usepackage{tikz}
\usepackage{graphicx}
\usepackage{amsmath}
\usepackage{caption, subcaption}

\title{Contrasting Human- and Machine-Generated Word-Level Adversarial Examples for Text Classification}

\author{
  \textbf{Maximilian Mozes}$^1$ \quad \textbf{Max Bartolo}$^1$ \quad \textbf{Pontus Stenetorp}$^1$ \quad \\
  \textbf{Bennett Kleinberg}$^{2,1}$ \quad \textbf{Lewis D. Griffin}$^1$ \\
  $^{1}$University College London \quad $^{2}$Tilburg University \\
  \small{\texttt{\{m.mozes, m.bartolo, p.stenetorp, l.griffin\}@cs.ucl.ac.uk}} \\
  \small{\texttt{bennett.kleinberg@tilburguniversity.edu}}
}

\begin{document}
\maketitle
\begin{abstract}
Research shows that natural language processing models are generally considered to be vulnerable to adversarial attacks; but recent work has drawn attention to the issue of validating these adversarial inputs against certain criteria (e.g., the preservation of semantics and grammaticality). Enforcing constraints to uphold such criteria may render attacks unsuccessful, raising the question of whether valid attacks are actually feasible. In this work, we investigate this through the lens of human language ability. We report on crowdsourcing studies in which we task humans with iteratively modifying words in an input text, while receiving immediate model feedback, with the aim of causing a sentiment classification model to misclassify the example. Our findings suggest that humans are capable of generating a substantial amount of adversarial examples using semantics-preserving word substitutions. We analyze how human-generated adversarial examples compare to the recently proposed \textsc{TextFooler}, \textsc{Genetic}, \textsc{BAE} and \textsc{SememePSO} attack algorithms on the dimensions naturalness, preservation of sentiment, grammaticality and substitution rate. Our findings suggest that human-generated adversarial examples are not more able than the best algorithms to generate natural-reading, sentiment-preserving examples, though they do so by being much more computationally efficient.
\end{abstract}

\section{Introduction}
The vulnerability of natural language processing (NLP) models to adversarial examples has received widespread attention~\cite{alzantot-etal-2018-generating, iyyer2018adversarial, ren-etal-2019-generating}. Text processing models have been shown to be susceptible to adversarial input perturbations across tasks, including question answering and text classification ~\cite{jia-liang-2017-adversarial, jin2019bert}. The concept of adversarial examples originated in computer vision~\cite{szegedy2013intriguing, goodfellow2014explaining}, and in that domain defines perturbations of input data to neural networks that are barely perceptible to the human viewer. Due to the discrete nature of text, however, that definition is less applicable in an NLP context, since every perturbation to the input tokens is unavoidably perceptible. Consequently, recent work aims to perturb textual inputs while preserving the sequence's naturalness and semantics (i.e., rendering changes imperceptible on these dimensions). However, as shown by~\citet{morris-etal-2020-reevaluating}, achieving these desiderata is challenging because even small perturbations can render a text meaningless, grammatically incorrect or unnatural, and furthermore several proposed adversarial attacks fail routinely to achieve them. If the algorithms are modified to ensure that they do achieve the desiderata then their rate of generating successful examples greatly diminishes, suggesting that the reported success rates of recently proposed attacks might represent an overestimation of their true capabilities. This, in turn, raises the question of whether valid word-level adversarial examples are routinely possible against trained NLP models.

\begin{figure*}[t]
    \centering
    \includegraphics[scale=0.28]{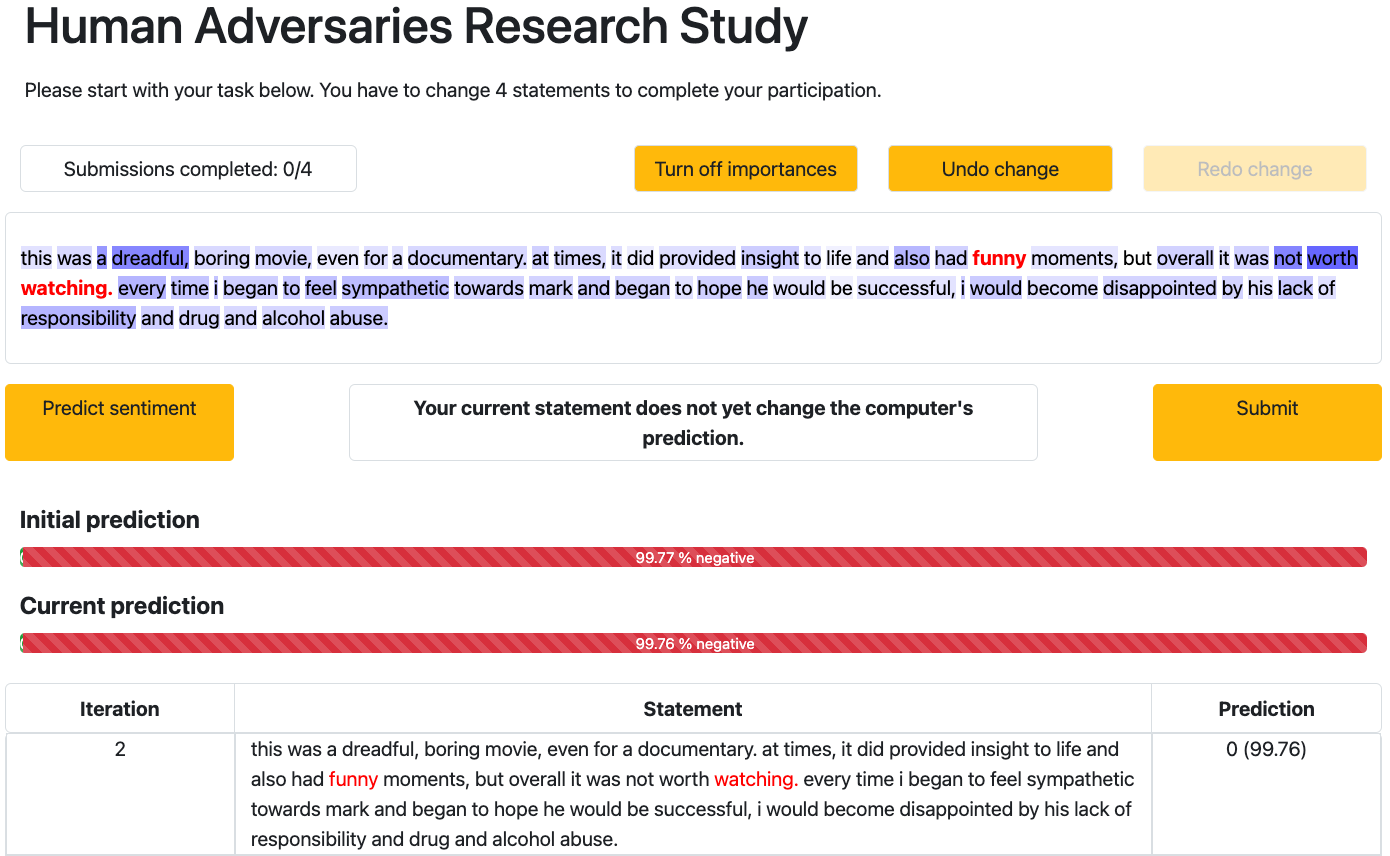}
    \caption{The interface for tasks 3 and 4. Participants are asked to change individual words in existing movie reviews to lead the RoBERTa model into misclassification. The word color highlighting represents the respective saliencies for each word in the sequence (see Section~\ref{subsec:stage_one} for details).}
    \label{fig:interface-task-3}
\end{figure*}

In this work, we aim to address this question by incorporating human judgments into the adversarial example generation process. Specifically, we report on a series of data collection efforts in which we task humans to generate adversarial examples from existing movie reviews, while instructed to strictly adhere to a set of validity constraints. In contrast to previous work~\cite[e.g.,][]{bartolo-etal-2020-beat, potts2020dynasent}, and in an attempt to replicate a word-level attack's mode of operation, human participants were only able to substitute individual words, and were not allowed to delete or insert new words into the sequence. This represents a black-box attack scenario, since human participants do not have access to information about the model's parameters or gradients. Participants worked in a web interface (Figure~\ref{fig:interface-task-3}) that allowed them to conduct word-level substitutions while receiving immediate feedback from a trained model.

After collecting the human-generated adversarial examples, we compare them to a set of automated adversarial examples for the same sequences using four recently proposed attacks: \textsc{TextFooler}~\cite{jin2019bert}, \textsc{Genetic}~\cite{alzantot-etal-2018-generating}, \textsc{BAE}~\cite{garg2020bae}, and \textsc{SememePSO}~\cite{zang-etal-2020-word}. Using human judgments from an independent set of crowdworkers, we assess for each generated adversarial example (human and automated) whether the perturbations changed the sequence's overall sentiment and whether they remained natural. 

We find that humans are capable of generating label-flipping word-level adversarial examples (i.e., the classifier misclassifies the sequence after human perturbation) in approximately 50\% of the cases. However, when comparing the ground truth labels of perturbed sequences to the sentiment labels provided by the independent set of human annotators, we find that only 58\% of the label-flipping human adversarial examples preserve their target sentiment after perturbation. This is considerably lower than for the best automated attacks, which exhibit a label consistency of up to 93\% (\textsc{TextFooler}) after perturbation. In terms of naturalness, we find no statistically significant differences between the human and machine attacks for the majority of comparisons. We furthermore observe that the human-generated sequences introduce fewer grammatical errors than most attacks.

These findings show that under similar constraints, machine-generated, word-level adversarial examples are comparable to human-generated ones with respect to their naturalness and grammaticality. Importantly, however, humans require, on average, only 10.9 queries to run against the model to generate label-flipping adversarial examples, while some attacks require thousands. We believe that our findings could further push the development of reliable word-level adversarial attacks in NLP, and our method and data might aid researchers in identifying human-inspired, more efficient ways of conducting adversarial word substitutions against neural text classification models.

The remainder of this paper is structured as follows. Section~\ref{sec:related_work} discusses previous work related to our research. Section~\ref{sec:human_generated_adv_examples} describes both phases of our data collection approach, i.e., the human generation of word-level adversarial examples and the subsequent validation of human- and machine-generated sequences with respect to their preservation of semantics and naturalness. This is followed by the analysis reported in Section~\ref{sec:analysis}, and a discussion of our findings and future work in Section~\ref{sec:discussion}. Finally, we conclude our paper in Section~\ref{sec:conclusion}.

\section{Related work}
\label{sec:related_work}
\paragraph{Adversarial attacks for NLP.}{
Adversarial attacks have been increasingly applied to NLP, with a diverse set of attack types being investigated, ranging from character-level edits~\citep{ebrahimi2017hotflip, ebrahimi-etal-2018-adversarial}, word-level replacements~\citep{alzantot-etal-2018-generating}, adding text to the input~\citep{jia-liang-2017-adversarial}, paraphrase-level modifications~\citep{iyyer2018adversarial, ribeiro-etal-2018-semantically}, to creating adversarial examples from scratch~\citep{bartolo-etal-2020-beat, nie2019adversarial}. This work focuses on word-level attacks.
}

\paragraph{Word-level attacks.}{
Our work builds on existing efforts on word-level adversarial attacks.
Attacks of this type can be further distinguished by whether the adversary has access to the model parameters (i.e., white-box) or is restricted to accessing only the predicted labels or confidence scores (i.e., black-box)~\citep{yuan2017adversarial}.
Word-level attacks have been explored for NLP tasks such as question answering~\citep{blohm-etal-2018-comparing, welbl-etal-2020-undersensitivity}, natural language inference~\citep{jin2019bert}, and text classification~\citep{papernot2016crafting, jin2019bert}.
A range of methodologies has been explored for finding optimal synonym substitutions, including population-based gradient-free optimization via genetic algorithms~\citep{alzantot-etal-2018-generating}, word saliency probability weighting~\citep{ren-etal-2019-generating}, similarity and consistency filtering~\citep{jin2019bert}, sememe-based word substitution and particle swarm optimization-based search~\citep{zang-etal-2020-word}, and contextual perturbations from masked language models~\citep{garg2020bae}.
Word-level perturbations have also been used as part of data augmentation strategies to certifiably improve model robustness~\citep{jia-etal-2019-certified}.

Existing efforts to detect and defend against word-level adversarial examples resort to adversarial data augmentation~\cite[e.g.,][]{ren-etal-2019-generating, jin2019bert} as well as rule-based~\cite{mozes-etal-2021-frequency} and learning-based~\cite{zhou-etal-2019-learning} approaches to identify adversarially perturbed inputs. 
}

\paragraph{Evaluating word-level attacks.}{
Of particular importance for this paper is how adversarial attacks can be evaluated against various dimensions. Adversarial attack performance has been shown to vary across evaluation dimensions including adversarial success rates, readability and content preservation~\citep{xu2020elephant}, as well as linguistic constraints such as semantics, grammaticality, the edit distance between original and perturbed text, and non-suspicion~\citep{morris-etal-2020-reevaluating}.
However, to the best of our knowledge, such evaluation efforts are limited to automated attacks and how humans perform at creating word-level adversarial examples across these dimensions remains unexplored.
}

\paragraph{Human-in-the-loop adversarial examples.}{
When the task is unconstrained, human crowdworkers have been shown to be capable of creating high quality adversarial examples for a variety of NLP tasks such as question answering~\citep{wallace2019trick, dua-etal-2019-drop, bartolo-etal-2020-beat, khashabi-etal-2020-bang}, natural language inference, \citep{nie2019adversarial}, and sentiment analysis~\citep{potts2020dynasent}.
We extend this line of work and investigate whether human capabilities for creating adversarial examples persist when the examples are constrained to arise from word-level perturbations, which have been shown to be highly effective~\cite{alzantot-etal-2018-generating, jin2019bert}.
}

\section{Method}
\label{sec:human_generated_adv_examples}
Our data collection process has two stages: first, we ask human annotators to perform a word-level adversarial attack for given input sequences. To this end, we prepared an online interface that lets participants perturb input sequences on a word-level whilst receiving immediate feedback as to how their changes affected classifier confidence. Second, we ask an independent set of crowdworkers to evaluate the generated adversarial examples.
 
\subsection{Stage one: human-generated word-level adversarial examples}
\label{subsec:stage_one}
In order to familiarize participants with the concept of word-level adversarial attacks for stage one of the data collection, we lead them through a sequence of four subtasks, each building on the preceding one:
\begin{enumerate} \itemsep -2pt
    \item Participants are asked to freely write a movie review with a specified sentiment
    \item Participants are asked to freely write an adversarial example
    \item Participants are given an existing movie review and are asked to use word-level adversarial perturbations \textit{without} adhering to semantic preservation and grammatical correctness
    \item Same as 3, but with the constraints to preserve semantics and grammatical correctness
\end{enumerate}
The data collected in tasks 1, 2 and 3 are not further analyzed in this paper, since these tasks were intended to help participants understand adversarial examples for text classification. After having successfully completed the three preparation tasks, the participants are considered fit to conduct task 4, which is the main topic of interest in this paper. 
For each subtask, we ask participants to submit four instances. For tasks 3 and 4, we randomly select four test set samples from the IMDb movie reviews dataset~\cite{Maas2011} for each participant. The reference classifier is a RoBERTa model~\cite{liu2019roberta} fine-tuned on IMDb, as it has been shown to perform highly on this task.\footnote{Specifically, we use a RoBERTA-base model provided by \texttt{HuggingFace}~\cite{Wolf2019HuggingFacesTS}, with 125 million parameters.} Our fine-tuned model achieves an accuracy of 93.8\% on the IMDb test set.\footnote{We randomly sample 1,000 training set sequences for epoch validation, and the final selected model achieves an accuracy of 92.7\% on this validation set.}

For tasks 1 and 2, the participants were able to directly see the classifier prediction before they submitted their reviews through clicking a button that queries the current sequence against the sentiment classification model.
For tasks 3 and 4, we asked participants to submit at least 15 iterations of word-level substitutions before moving on to the next review.\footnote{We tested the task with different numbers of iterations, and found this number to be suitable for our experiments.} After each submitted iteration the model provided immediate feedback as to how the change affected its prediction. The sequence of display of the four reviews in tasks 3 and 4 is based on the review length in ascending order. 

\paragraph{Word saliencies.}{
For tasks 3 and 4, the interface additionally displays the word saliencies~\cite{li-etal-2016-visualizing, word-saliency-li} for each word in the movie review. Here, the word saliency is defined as the model's difference in prediction confidence before and after replacing the word with an out-of-vocabulary token. The interface for tasks 3 and 4 is shown in Figure~\ref{fig:interface-task-3}.\footnote{Participants were given the option to disable the word saliency highlighting, and were also able to undo and redo changes made to the input sequence.}
}

We use Amazon's Mechanical Turk to collect the data. We restrict participation to workers that have previously conducted more than 1,000 successful Human Intelligence Tasks (HITs), have an approval rate of above 98\% and who are located in Canada, the US, or the UK. We estimate the completion time to be under 60 minutes, and pay USD 12.40 per user per HIT. In total, we collected responses from $n=43$ participants. For task 4, we had to exclude two individual submissions due to technical errors. The resulting sample consists of 172 collected reviews for the first three tasks and 170 reviews for task 4. Despite a random allocation of test set sequences to participants, we did not encounter duplicate sequences in the sample.\footnote{The data are available at \url{http://github.com/maximilianmozes/human_adversaries}.}

\paragraph{Comparison to automated attacks.}{
We compare the human-generated, word-level adversarial examples against a set of automatically generated ones. Specifically, we attack the fine-tuned RoBERTa model as used for the data collection phase on the 170 sequences collected in task 4. We experiment with four recently proposed attacks.
}

\paragraph{\textsc{Genetic}.}{
The \textsc{Genetic} attack~\cite{alzantot-etal-2018-generating} uses a population-based genetic search method to generate word-level adversarial examples. Specifically, the attack iteratively adds individual perturbations to an input sequence until the model misclassifies the perturbed input.
}

\paragraph{\textsc{TextFooler}.}{
\textsc{TextFooler}~\cite{jin2019bert} is a black-box word-level adversarial attack that ranks words according to their importance for classifier decision-making, and then iteratively replaces the selected words with semantically similar ones to lead the model into misclassification. \textsc{TextFooler} ensures that the replacement tokens have the same part-of-speech as the selected word. Furthermore, the algorithm utilizes the Universal Sentence Encoder~\cite{cer2018universal} to identify replacements that best preserve sequence semantics. 
}

\paragraph{\textsc{SememePSO}.}{
Whereas existing work predominantly relies on embedding spaces or thesauri like WordNet~\cite{fellbaum98wordnet},~\citet{zang-etal-2020-word} propose an attack using sememes (which the authors describe as minimum semantic units of language) to identify semantics-preserving word substitutions. The attack, referred to as \textsc{SememePSO}, additionally uses a combinatorial optimization method based on particle swarm optimization.
}

\paragraph{\textsc{BAE}.}{
In contrast to previous approaches,~\citet{garg2020bae} propose BERT-based Adversarial Examples (\textsc{BAE}), an attack that relies on a BERT masked language model used to both replace and insert new tokens into an existing sequence to generate an adversarial example. They introduce multiple variants of \textsc{BAE} and in this work, we experiment with the \textsc{BAE-R} variant, which only replaces tokens, but does not insert new ones. This is to ensure that \textsc{BAE} is directly comparable to the other attacks analyzed in our experiments.
}

We generate adversarial examples based on the 170 sequences used during the data collection study, and use the \texttt{TextAttack}~\cite{morris-etal-2020-textattack} Python library with all attacks in their default configuration. For computational efficiency, for the \textsc{Genetic} attack, we use a slightly different variant compared to~\citet{alzantot-etal-2018-generating}. Specifically, we use the \texttt{faster-alzantot} variant offered by \texttt{TextAttack}, which implements the modifications suggested in~\citet{jia-etal-2019-certified}.

\subsection{Stage two: evaluating generated adversarial examples}
To evaluate the adversarial examples generated by algorithmic approaches and human participants in stage one, we ask an independent set of crowdworkers to annotate the collected data. Specifically, in a new data collection stage, participants read and judged each adversarial example on its sentiment and naturalness, both on a five-point Likert scale. Here, a rating of 1 would denote very negative sentiment (a very unnatural review), whereas a rating of 5 would indicate a very positive sentiment (a very natural review). We use the sentiment judgments to measure the deviation of sentiment resulting from introducing the perturbations (high deviations imply a larger shift in sentiment), and the naturalness judgment to evaluate whether the adversarial substitutions distort the naturalness of the sequence. Specifically, we ask participants to rate the 172 generated adversarial examples from task 2, the 170 unperturbed reviews used in task 4, and the corresponding human- and machine-generated adversarial examples. For the examples in task 4, we select the first label-flipping iteration for a successful submission, and the iteration which exhibits the lowest confidence on the ground truth for unsuccessful submissions.

We recruited participants via the Prolific Academic\footnote{\url{https://www.prolific.co/}} platform, and aimed to collect three independent ratings per text. We used independent workers per criterion and recruited 120 participants for each. Each participant was asked to rate 30 texts (randomly selected from all available sequences) and received GBP 1.50 as compensation. On average, each text was rated by 3.55 human judges.

\begin{table}[t]
\centering
\resizebox{\columnwidth}{!}{
\begin{tabular}{lrrr}
\toprule
\multicolumn{1}{c}{\textbf{Attack}} &
\multicolumn{1}{c}{\textbf{ASR}} &
\multicolumn{1}{c}{\textbf{Reference}} &
\multicolumn{1}{c}{\texttt{TextAttack}} \\
\midrule
\textsc{Human} & 48.8 & --- & --- \\
\textsc{Genetic} & 38.2 & 42.9 & 46.7 \\
\textsc{TextFooler} & 99.4 & 98.8 & 100.0 \\
\textsc{BAE} & 43.0 & 42.3 & 55.6 \\
\textsc{SememePSO} & 100.0 & 100.0 & 100.0 \\
\bottomrule
\end{tabular}
}
\caption{Attack success rates (ASR) on the 170 test set sequences. Reference denotes the success rate against an independent fine-tuned RoBERTa model, \texttt{TextAttack} refers to the success rates reported by~\citet{morris-etal-2020-textattack} against a BERT-Base model using 100 random sequences from IMDb.}
\label{tab:attack-success-rates}
\end{table}

\section{Analysis}
\label{sec:analysis}
After collecting the human judgments we analyze both the human and machine attacks' performance on generating adversarial examples. The primary objective is to investigate the feasibility of word-level adversarial examples that adhere to validity criteria as suggested in previous work~\cite{morris-etal-2020-reevaluating}. We use the \textit{attack success rate} (ASR) as the initial metric to evaluate the performance of either attack mode (human and algorithmic). The attack success rate is defined as the percentage of successful adversarial examples (i.e., those that are misclassified after perturbation) to all perturbed sequences.

\begin{figure}[t]
    \resizebox{1.0\columnwidth}{!}{
    \centering
    \includegraphics[scale=0.28]{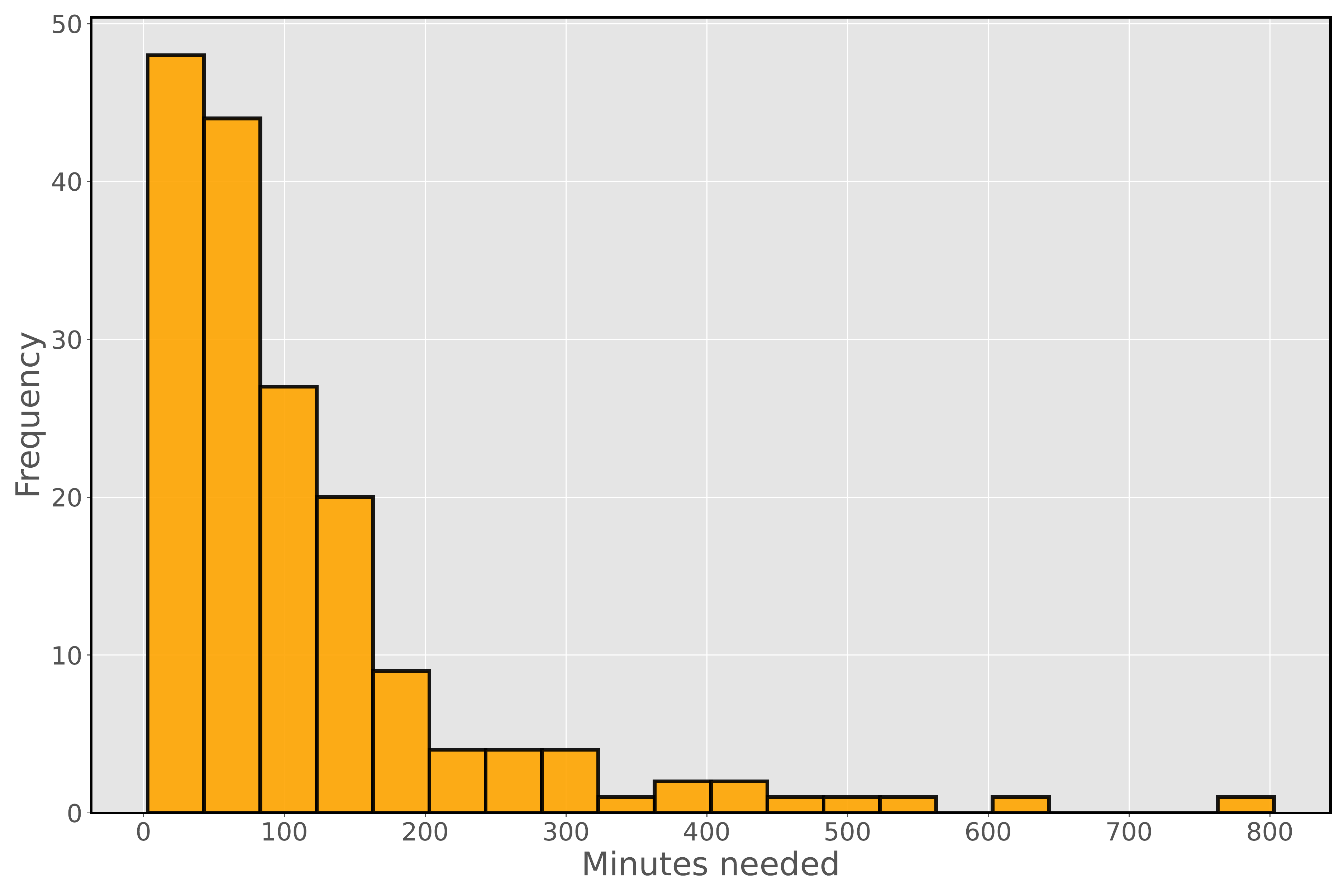}
    }
    \caption{Minutes needed by participants for task 4.}
    \label{fig:task-4-times-needed}
\end{figure}

We observe that overall, workers were generally able to generate successful movie reviews for task 1 (for 90\% of the submitted sequences the model predicted the desired sentiment) and led the model into misclassification in task 2 for the majority of the cases (ASR 80\%). For task 3, workers also managed to flip the model prediction by introducing arbitrary word-level perturbations (ASR 86\%). Crucially, when we introduced constraints in task 4, the ASR drops to 49\%, suggesting an increased difficulty of generating word-level adversarial examples when attempting to preserve the sentiment and naturalness of the text. It is worth mentioning that we conducted additional experiments with expert annotators (i.e., academic researchers with experience in NLP) and found that the ASR for task 4 was even lower compared to the crowdworkers. As a comparison, we report the ASR of all word-level attacks in Table~\ref{tab:attack-success-rates}, and observe that the \textsc{Human} ASR is higher than the ones for \textsc{Genetic} and \textsc{BAE}, but lower than \textsc{TextFooler} and \textsc{SememePSO}.

Figure~\ref{fig:task-4-times-needed} depicts the distribution of times needed for the human participants to generated the word-level adversarial examples in task 4. We observe that participants need on average 111.29 minutes (standard deviation: 119.77) to complete the task. 

\begin{table}[t]
\centering
\resizebox{\columnwidth}{!}{
\begin{tabular}{lcccc}
\toprule
\multicolumn{1}{c}{\textbf{Attack}} &
\multicolumn{1}{c}{\textbf{Match (S)}} &
\multicolumn{1}{c}{$\Delta_S$} &
\multicolumn{1}{c}{\textbf{Match (U)}} &
\multicolumn{1}{c}{$\Delta_U$} \\
\midrule
\textsc{Human} & 58\% & 1.15 (1.10) & 90\% & 0.35 (0.82) \\
\textsc{Genetic} & 86\% & 0.33 (0.85) &  98\% & 0.23 (0.65) \\
\textsc{TextFooler} & 93\% & 0.28 (0.68) & 100\% & 0.60 (0.00) \\
\textsc{BAE} & 82\% & 0.29 (0.88) & 97\% & 0.29 (0.52) \\
\textsc{SememePSO} & 82\% & 0.47 (0.89) & --- & --- \\
\bottomrule
\end{tabular}
}
\caption{The percentage of sentiment-preserving adversarial examples per attack. Match (S) denotes the percentage of label-flipping (successful) samples that preserve sentiment, Match (U) denotes unsuccessful ones.}
\label{tab:label-matches-sentiment}
\end{table}

\begin{table*}[t]
\centering
\resizebox{\textwidth}{!}{
\begin{tabular}{lccccc}
\toprule
& \textsc{Human} & \textsc{Genetic} & \textsc{BAE} & \textsc{TextFooler} & \textsc{SememePSO} \\
\midrule
\textsc{Human} & --- & $-0.32\,\,[-0.79; 0.16]$ & $-0.98\,\,[-1.49; -0.47]$* & --- & --- \\
\textsc{Genetic} & $-0.55\,\,[-1.22; 0.12]$ & --- & $-0.63\,\,[-1.09; -0.17]$* & --- & --- \\
\textsc{BAE} & $-0.23\,\,[-0.88; 0.42]$ & $0.29\,\,[-0.33; 0.91]$ & --- & --- & --- \\
\textsc{TextFooler} & $0.13\,\,[-0.41; 0.67]$ & $0.65\,\,[0.13; 1.16]$* & $0.35\,\,[-0.14; 0.85]$ & --- & --- \\
\textsc{SememePSO} & $-0.26\,\,[-0.81; 0.30]$ & $0.27\,\,[-0.25; 0.79]$ & $-0.02\,\,[-0.53; 0.48]$ & $-0.38\,\,[-0.76; -0.01]$* & --- \\
\bottomrule
\end{tabular}
}
\caption{Cohen's $d$ effect sizes for naturalness comparisons. The lower triangle represents comparisons for successful adversarial examples, the upper one those for unsuccessful examples. The table can be read row-wise, such that the rating differences are computed by subtracting the mean of the column attack from the mean of the row attack (i.e., a negative effect size indicates that the mean naturalness difference of the row attack is lower than that of the column attack). * denotes statistically significant differences. }
\label{tab:statistical-results-naturalness}
\end{table*}

\subsection{Analysis of human annotations}
\paragraph{Sentiment.}{
\label{subsec:sentiment-analysis}
We define the final sentiment value for each text as negative if its mean rating is below 3.0, and positive if above.\footnote{80 samples with a mean rating of exactly 3.0 were excluded from our analysis.} As an initial test, we compute the correlation between the ground truth label (positive or negative) and the mean human sentiment rating for unperturbed samples for task 4. We obtain a Pearson correlation of $r=0.89$ ($95\%\,\,\mathrm{CI}=[0.85, 0.92]$, $p < .001$). This demonstrates high agreement between the IMDb ground truth labels and the human annotations for both tasks.

Next, we want to assess whether adversarial examples preserve the sentiment of the original sequence. To test this, we compare the ground truth label for each text with its binarized human sentiment label and consider sentiment to have been preserved when these agree. Table~\ref{tab:label-matches-sentiment} shows the proportion of adversarial examples whose ground truth label matches the binarized human rating. $\Delta_S$ and $\Delta_U$ represent the mean (standard deviation) differences in ratings between the original and adversarial sequences. The higher the difference, the more do human ratings between the unperturbed and perturbed sequences deviate from each other.

All algorithmic attacks show high values (above 80\%) for successful examples, while the \textsc{Human} attacks preserve the sentiment less often (58\%). Similarly, the mean distance ($\Delta_S=1.15$) for the \textsc{Human} attack is considerably higher than that for the algorithmic attacks. Thus, of the human-generated adversarial examples, only 58\% preserve the original sentiment and can be considered for further evaluation. The central question now is whether the higher sentiment-preservation rate of algorithmic attacks holds up if we submit the data to a naturalness test.

\begin{table}[t] 
\centering
\footnotesize{
\begin{tabular}{lccc}
\toprule
\multicolumn{1}{c}{\textbf{Attack}} &
\multicolumn{1}{c}{$\Delta_S$} &
\multicolumn{1}{c}{$\Delta_U$} &
\multicolumn{1}{c}{$\Delta_{comb}$} \\
\midrule
\textsc{Human} & 0.50 (1.25) & 0.14 (1.33) & 0.27 (1.31)\\
\textsc{Genetic} & -0.16 (1.16) & 0.55 (1.29) & 0.32 (1.29) \\
\textsc{TextFooler} & 0.67 (1.32) & 2.67 (0.00) & 0.68 (1.33) \\
\textsc{BAE} & 0.20 (1.33) & 1.30 (1.05) & 0.89 (1.27) \\
\textsc{SememePSO} & 0.17 (1.28) & --- & 0.17 (1.28) \\
\bottomrule
\end{tabular}
}
\caption{The differences (mean and standard deviation) between the average naturalness rating for the unperturbed and attacked sequences for successful ($\Delta_S$) and unsuccessful ($\Delta_U$) adversarial examples as well as their combination ($\Delta_{comb})$. Positive values indicate a decrease in naturalness. Histograms highlighting the distribution of mean ratings can be found in Figure~\ref{fig:naturalness-visualization} of the Appendix.}
\label{tab:rating-differences-naturalness}
\end{table}
}

\paragraph{Naturalness.}{
Similar to sentiment, we now compare the naturalness ratings between the unperturbed and attacked sequences. The average naturalness rating per text is compared between unperturbed texts and their adversarial counterparts. The larger that difference, the more unnatural the adversarial perturbations have rendered the respective movie review. We only consider the sentiment-preserving adversarial examples as explained in Section~\ref{subsec:sentiment-analysis}.

To test statistically, whether the attacks differed in their naturalness deviation, we ran a 5 (\textit{attack types}) by 2 (\textit{success:} successful and unsuccessful) ANOVA with the naturalness differences as the dependent variable. That analysis yielded a significant main effect of attack type, $F(4, 666) = 7.87, p < 0.001$ and success, $F(1, 666) = 18.64, p < 0.001$, both of which were subsumed in the interaction effect, $F(3, 666) = 7.29, p < 0.001$.

To disentangle the interaction effect, we show the Cohen's $d$ effect sizes~\cite{cohen2013statistical} for the attack type comparisons for successful and unsuccessful attacks. This analysis helps us to understand how the effect of attack type depends on the attack's success. The effect size $d$ expresses the absolute magnitude of the mean naturalness difference per comparison and is preferred over $p$-values.\footnote{$d= 0.2$, $d= 0.5$ and $d= 0.8$ can be interpreted as a small, medium and large effects, respectively.} Table~\ref{tab:statistical-results-naturalness} shows the $d$ values with their 99.75\% ($p=0.05/20$) confidence intervals (CI). A CI containing zero implies that the difference in naturalness cannot be considered statistically significant and therefore be disregarded. For the unsuccessful examples, the comparisons are missing for the \textsc{TextFooler} and \textsc{SememePSO} attacks. This is because both attacks are highly successful, such that only a single (\textsc{TextFooler}) and none (\textsc{SememePSO}) of the adversarial examples did not flip the classifier's prediction.

No differences emerge between the mean naturalness rating difference for the majority of comparisons with respect to the \textsc{Human} attack. Only for the unsuccessful adversarial examples do we see that the rating differences between \textsc{Human} and \textsc{BAE} are significantly different. As a whole, this analysis suggests that in terms of naturalness, the \textsc{Human} adversarial examples are not significantly different from the machine-generated ones (see Table~\ref{tab:rating-differences-naturalness} for the means). 

\begin{table*}[t]
\centering
\resizebox{0.85\textwidth}{!}{
\begin{tabular}{lcccc}
\toprule
\multicolumn{1}{c}{\textbf{Attack}} &
\multicolumn{1}{c}{$\mathrm{Sub}_{S}$} &
\multicolumn{1}{c}{$\mathrm{Sub}_{U}$} &
\multicolumn{1}{c}{$\mathrm{Q}_{S}$} &
\multicolumn{1}{c}{$\mathrm{Q}_{U}$} \\
\midrule
\textsc{Human} & 7.5 (9.2) & 8.6 (8.9)$^a$ & 10.9 (13.8) & 17.5 (10.7) \\
\textsc{Genetic} & 6.9 (4.2)$^d$ & 14.0 (4.8)$^{c,d}$ & 3558.1 (2102.5) & 8069.1 (1211.4) \\
\textsc{TextFooler} & 8.4 (8.0)$^{d,e}$ & 40.3 (0.0) & 515.2 (379.3) & 1821.0 (0.0) \\
\textsc{BAE} & 4.0 (2.9)$^{a,b}$ & 9.6 (1.4)$^a$ & 292.8 (112.3) & 435.8 (149.4) \\
\textsc{SememePSO} & 5.4 (4.1)$^b$ & --- & 140956.3 (148494.5) & --- \\
\bottomrule
\end{tabular}
}
\caption{Mean (SD) substitution rates ($\mathrm{Sub}$) and the number of queries ($Q$) per attack on all sentiment-preserving adversarial examples. Subscripts $S$ and $U$ denote label-flipping and unsuccessful attacks, respectively. Superscripts indicate significant differences with $^a$\textsc{Genetic}, $^b$\textsc{TextFooler}, $^c$\textsc{Human}, $^d$\textsc{BAE}, and $^e$\textsc{SememePSO} attacks.}
\label{tab:sub-rate-naturalness}
\end{table*}

\begin{table}[t]
\centering
\footnotesize{
\begin{tabular}{lcc}
\toprule
\multicolumn{1}{c}{\textbf{Attack}} &
\multicolumn{1}{c}{\textbf{Num. errors}} &
\multicolumn{1}{c}{\textbf{Adv. errors} (\%)} \\
\midrule
None & 10.8 (5.7)$^\ast$ & --- \\
\textsc{Human} & 11.2 (5.6)$^\ast$ & 34.7 \\
\textsc{Genetic} & 11.1 (5.7)$^\ast$ & 37.1 \\
\textsc{TextFooler} & 11.7 (5.7)$^\ast$ & 56.5 \\
\textsc{BAE} & 15.0 (6.1)$^\ast$ & 92.4 \\
\textsc{SememePSO} & 11.0 (5.8)$^\ast$ & 22.4 \\
\bottomrule
\end{tabular}
}
\caption{Mean (SD) number of errors made per attack and the percentage of cases in which the adversarial example contains more grammatical errors than its unperturbed counterpart (Adv. errors). None represents the unperturbed reviews. $^\ast$indicates significant difference with \textsc{BAE}.}
\label{tab:grammatical-comparison}
\end{table}

\begin{table*}[t]
\centering
\footnotesize
\begin{tabular}{{l p{0.49\textwidth} c c c }}
\toprule
\textbf{Attack} & \textbf{Text} & \textbf{Pred.} & \textbf{Naturalness} & \textbf{Sentiment} \\
\midrule
--- & it boggles the mind how big name stars such as those in this movie can be part of the one of the dullest movies i ve ever seen. & \textit{negative} & 4.5 & 1.9 \\
\textsc{Human} & it \color{red}\textbf{amazes}\color{black}\, the mind how big name stars such as those in this movie can be part of the one of the \color{red}\textbf{simplest}\color{black}\, movies i ve ever seen.
 & \textit{positive} & 4.3 & 1.4 \\
\textsc{Genetic} & it boggles the mind how big \color{red}\textbf{naming}\color{black}\, stars such as those in this movie can be part of the one of the dullest \color{red}\textbf{cinema}\color{black}\, i ve \color{red}\textbf{always}\color{black}\, \color{red}\textbf{observed.}\color{black}\,
 & \textit{negative} & 1.5 & 1.8 \\
\textsc{BAE} & it boggles the mind how big name stars such as those in this movie can be part of the one of the \color{red}\textbf{liest}\color{black}\, movies i ve ever seen.
 & \textit{positive} & 3.7 & 1.0 \\
\textsc{TextFooler} & it boggles the mind how big name stars such as those in this movie can be part of the one of the \color{red}\textbf{neatest}\color{black}\, movies i ve ever seen.
 & \textit{positive} & 4.0 & 1.0 \\
\textsc{SememePSO} & it boggles the mind how big name stars such as those in this movie can be part of the one of the \color{red}\textbf{deepest}\color{black}\, movies i ve ever seen.
 & \textit{positive} & 4.3 & 1.0 \\
\bottomrule
\end{tabular}
\caption{An example movie review from IMDb together with its corresponding adversarial examples. The Naturalness and Sentiment columns denote the mean ratings as explained in Section~\ref{subsec:sentiment-analysis}. Individual examples have been reduced to excerpts for better readability, the full texts can be found in Table~\ref{fig:illustrations-adversarial-examples-full} of the Appendix.}
\label{fig:illustrations-adversarial-examples}
\end{table*}
}

\subsection{Substitution rate and number of queries}
Next, we analyze the effect of the substitution rate for each adversarial example on its corresponding naturalness rating as well as the number of model queries required per attack. Statistical testing with an ANOVA showed that there were significant main effects of attack type and success as well a significant interaction. Table~\ref{tab:sub-rate-naturalness} indicates significant differences between the comparisons. Further, we observe a negative Pearson correlation of $r=-0.31$ ($95\%\,\,\mathrm{CI}=[-0.38, -0.24]$, $p < .001$) between the mean naturalness ratings and the word substitution rate, indicating that the naturalness deteriorated with increasing  substitutions. Moreover, Table~\ref{tab:sub-rate-naturalness} shows that the automated attacks perform notably more model queries as compared to the \textsc{Human} attack.\footnote{Note that we do not consider the model queries used for computing the word saliencies provided to the crowdworkers in this comparison.} While some attacks query a model thousands of times for a single adversarial example, humans are able to find successful adversarial examples with an average of 10.9 queries run against a model. This suggests that humans are considerably more efficient in generating valid word-level adversarial examples. Together, these findings raise the question of how automated attacks might be further optimized with respect to their computational efficiency.

\subsection{Grammaticality}
As a last evaluation dimension, we look at the number of grammatical mistakes made between the original reviews and their adversarial counterparts. We follow~\citet{morris-etal-2020-reevaluating} by using the \texttt{LanguageTool}\footnote{\url{https://github.com/jxmorris12/language_tool_python}} grammar checker but exclude all errors related to the category \texttt{CASING} since all sequences have been lower-cased. We compare the mean number of grammatical errors made per attack and the percentage of unperturbed-adversarial sequence pairs for which the adversarial example has more grammatical errors than the unperturbed sequence. For the former, we conduct an ANOVA and compute effect sizes analogously to aforementioned experiments.

Table~\ref{tab:grammatical-comparison} suggests that all attacks produce texts with a higher number of grammatical errors than the unperturbed sequences. Among the different attacks, \textsc{BAE} generates considerably more grammatical errors (15.0 errors per review) than the other attacks (between 11.0 and 11.7 errors per review). The \textsc{SememePSO} attack has the lowest rate (22.4\%) of increasing grammatical errors. For 34.7\% of all tested sequences, the \textsc{Human} adversarial word substitutions yielded an increase in grammatical errors. The percentages of 37.1\% for the \textsc{Genetic} and 56.5\% for \textsc{TextFooler} are comparable to the results reported in~\citet{morris-etal-2020-reevaluating}.

Table~\ref{fig:illustrations-adversarial-examples} shows an example movie review from IMDb as well as the perturbed counterparts resulting from all attacks.

\section{Discussion}
\label{sec:discussion}
Despite some reported successes, recent work questions the validity of machine-generated word-level adversarial examples. Central to that critical view are evaluation criteria on which the adversarial examples fall short~\cite{morris-etal-2020-reevaluating}. The argument is that with these criteria as constraints, most (if not all) word-level adversarial examples are deemed invalid. In this work, we investigated how feasible such adversarial examples can be generated by humans when explicitly asked to respect a set of validity constraints. The underlying reasoning was that human performance might have been able to improve the quality standard of word-level adversarial examples.

Our findings suggest that with respect to the success rate as well as the preservation of semantics and naturalness, humans do not outperform state-of-the-art attack algorithms in generating word-level adversarial substitutions. But they also do not differ much. This finding speaks to the difficulty of the task. However, our findings suggest that while humans do not outperform machines with respect to the aforementioned criteria, they are able to generate adversarial examples of similar quality using a fraction of the attack iterations required by the algorithms. Humans are able to generate label-flipping examples with only a handful of queries, while the algorithmic attacks might need thousands of inference steps to find successful word substitutions. Further, humans do this without introducing more grammatical errors than the algorithmic attacks. In sum, this work suggests that humans produce adversarial examples comparable to state-of-the-art attacks but at a fraction of the computational costs. With a better understanding of how humans achieve this, future work could try to close that gap and develop more computationally efficient algorithmic adversarial attacks inspired by human language reasoning.

\subsection{Limitations and future work}
Our work comes with various limitations. First, the broad distribution of human naturalness ratings of unperturbed IMDb test set sequences reflects the informal style of these texts. Future work would need to assess whether our results would differ in more formal writing (e.g., journalistic or academic writing) where finding adequate replacements while meeting the quality criteria could be even harder. Second, with respect to the number of queries, a direct comparison between the success rates of human and algorithmic attacks might be misleading, since asking humans to conduct thousands of iterations per sequence is practically infeasible. Future work could assess how algorithmic attacks perform if constrained to the same number of iterations as humans.

Moreover, the notable difference in efficiency between humans and algorithms needs to be investigated further, for example by analyzing human strategies in conducting word substitutions, which can potentially be beneficial for developing more efficient attack algorithms. 

Additionally, our findings support previous work~\cite{morris-etal-2020-reevaluating} and suggest that word-level adversarial attacks might impose unrealistic constraints (even on humans). This observation raises the question of whether an attention shift towards phrase-based adversarial examples is needed to guarantee the validity of adversarial examples in NLP. To this end, it would be interesting to expand our research focus beyond word-level attacks, for example by relaxing the constraint on word-level substitutions for humans and giving them additional degrees of freedom to rephrase sequences in individual iterations.

\section{Conclusion}
\label{sec:conclusion}
This paper compared human and machine performance on generating word-level adversarial examples against a text classification model for sentiment analysis. We observe that human-generated adversarial examples do not preserve a sequence's sentiment as well as machine-generated ones do, but are similar in terms of their naturalness after label-flipping perturbation. While these findings do not suggest that humans outperform algorithms for this task, we find that they achieve similar performance in a much more efficient manner. We therefore believe that our work can build the foundation for future research aiming to further optimize algorithmic word-level attacks by potentially adapting human-inspired strategies for this task.

\section*{Acknowledgements}
This research was supported by the Dawes Centre for Future Crime at University College London.

\section*{Ethical considerations}
This work uses publicly available data ~\cite{Maas2011} and data collected from human participants. All human participants provided informed consent and the studies were approved by the local ethics review board. No personal information was collected.

\bibliography{anthology,custom}

\begin{thebibliography}{36}
\expandafter\ifx\csname natexlab\endcsname\relax\def\natexlab#1{#1}\fi

\bibitem[{Alzantot et~al.(2018)Alzantot, Sharma, Elgohary, Ho, Srivastava, and
  Chang}]{alzantot-etal-2018-generating}
Moustafa Alzantot, Yash Sharma, Ahmed Elgohary, Bo-Jhang Ho, Mani Srivastava,
  and Kai-Wei Chang. 2018.
\newblock \href {https://doi.org/10.18653/v1/D18-1316} {Generating natural
  language adversarial examples}.
\newblock In \emph{Proceedings of the 2018 Conference on Empirical Methods in
  Natural Language Processing}, pages 2890--2896, Brussels, Belgium.
  Association for Computational Linguistics.

\bibitem[{Bartolo et~al.(2020)Bartolo, Roberts, Welbl, Riedel, and
  Stenetorp}]{bartolo-etal-2020-beat}
Max Bartolo, Alastair Roberts, Johannes Welbl, Sebastian Riedel, and Pontus
  Stenetorp. 2020.
\newblock \href {https://doi.org/10.1162/tacl_a_00338} {Beat the {AI}:
  Investigating adversarial human annotation for reading comprehension}.
\newblock \emph{Transactions of the Association for Computational Linguistics},
  8:662--678.

\bibitem[{Blohm et~al.(2018)Blohm, Jagfeld, Sood, Yu, and
  Vu}]{blohm-etal-2018-comparing}
Matthias Blohm, Glorianna Jagfeld, Ekta Sood, Xiang Yu, and Ngoc~Thang Vu.
  2018.
\newblock \href {https://doi.org/10.18653/v1/K18-1011} {Comparing
  attention-based convolutional and recurrent neural networks: Success and
  limitations in machine reading comprehension}.
\newblock In \emph{Proceedings of the 22nd Conference on Computational Natural
  Language Learning}, pages 108--118, Brussels, Belgium. Association for
  Computational Linguistics.

\bibitem[{Cer et~al.(2018)Cer, Yang, Kong, Hua, Limtiaco, John, Constant,
  Guajardo-C{\'e}spedes, Yuan, Tar et~al.}]{cer2018universal}
Daniel Cer, Yinfei Yang, Sheng-yi Kong, Nan Hua, Nicole Limtiaco, Rhomni~St
  John, Noah Constant, Mario Guajardo-C{\'e}spedes, Steve Yuan, Chris Tar,
  et~al. 2018.
\newblock Universal sentence encoder.
\newblock \emph{arXiv preprint arXiv:1803.11175}.

\bibitem[{Cohen(1988)}]{cohen2013statistical}
Jacob Cohen. 1988.
\newblock \href
  {https://books.google.co.uk/books?id=2v9zDAsLvA0C&pg=PP1&redir_esc=y#v=onepage&q&f=false}
  {\emph{Statistical power analysis for the behavioral sciences}}.
\newblock Academic press.

\bibitem[{Dua et~al.(2019)Dua, Wang, Dasigi, Stanovsky, Singh, and
  Gardner}]{dua-etal-2019-drop}
Dheeru Dua, Yizhong Wang, Pradeep Dasigi, Gabriel Stanovsky, Sameer Singh, and
  Matt Gardner. 2019.
\newblock \href {https://doi.org/10.18653/v1/N19-1246} {{DROP}: A reading
  comprehension benchmark requiring discrete reasoning over paragraphs}.
\newblock In \emph{Proceedings of the 2019 Conference of the North {A}merican
  Chapter of the Association for Computational Linguistics: Human Language
  Technologies, Volume 1 (Long and Short Papers)}, pages 2368--2378,
  Minneapolis, Minnesota. Association for Computational Linguistics.

\bibitem[{Ebrahimi et~al.(2018{\natexlab{a}})Ebrahimi, Lowd, and
  Dou}]{ebrahimi-etal-2018-adversarial}
J.~Ebrahimi, Daniel Lowd, and D.~Dou. 2018{\natexlab{a}}.
\newblock On adversarial examples for character-level neural machine
  translation.
\newblock In \emph{COLING}.

\bibitem[{Ebrahimi et~al.(2018{\natexlab{b}})Ebrahimi, Rao, Lowd, and
  Dou}]{ebrahimi2017hotflip}
Javid Ebrahimi, Anyi Rao, Daniel Lowd, and Dejing Dou. 2018{\natexlab{b}}.
\newblock \href {https://doi.org/10.18653/v1/P18-2006} {{H}ot{F}lip: White-box
  adversarial examples for text classification}.
\newblock In \emph{Proceedings of the 56th Annual Meeting of the Association
  for Computational Linguistics (Volume 2: Short Papers)}, pages 31--36,
  Melbourne, Australia. Association for Computational Linguistics.

\bibitem[{Fellbaum(1998)}]{fellbaum98wordnet}
Christiane Fellbaum, editor. 1998.
\newblock \href {https://mitpress.mit.edu/books/wordnet} {\emph{{WordNet: an
  electronic lexical database}}}.
\newblock MIT Press.

\bibitem[{Garg and Ramakrishnan(2020)}]{garg2020bae}
Siddhant Garg and Goutham Ramakrishnan. 2020.
\newblock Bae: Bert-based adversarial examples for text classification.
\newblock \emph{arXiv preprint arXiv:2004.01970}.

\bibitem[{Goodfellow et~al.(2014)Goodfellow, Shlens, and
  Szegedy}]{goodfellow2014explaining}
Ian~J Goodfellow, Jonathon Shlens, and Christian Szegedy. 2014.
\newblock Explaining and harnessing adversarial examples.
\newblock \emph{arXiv preprint arXiv:1412.6572}.

\bibitem[{Iyyer et~al.(2018)Iyyer, Wieting, Gimpel, and
  Zettlemoyer}]{iyyer2018adversarial}
Mohit Iyyer, John Wieting, Kevin Gimpel, and Luke Zettlemoyer. 2018.
\newblock Adversarial example generation with syntactically controlled
  paraphrase networks.
\newblock \emph{arXiv preprint arXiv:1804.06059}.

\bibitem[{Jia and Liang(2017)}]{jia-liang-2017-adversarial}
Robin Jia and Percy Liang. 2017.
\newblock \href {https://doi.org/10.18653/v1/D17-1215} {Adversarial examples
  for evaluating reading comprehension systems}.
\newblock In \emph{Proceedings of the 2017 Conference on Empirical Methods in
  Natural Language Processing}, pages 2021--2031, Copenhagen, Denmark.
  Association for Computational Linguistics.

\bibitem[{Jia et~al.(2019)Jia, Raghunathan, G{\"o}ksel, and
  Liang}]{jia-etal-2019-certified}
Robin Jia, Aditi Raghunathan, Kerem G{\"o}ksel, and Percy Liang. 2019.
\newblock \href {https://doi.org/10.18653/v1/D19-1423} {Certified robustness to
  adversarial word substitutions}.
\newblock In \emph{Proceedings of the 2019 Conference on Empirical Methods in
  Natural Language Processing and the 9th International Joint Conference on
  Natural Language Processing (EMNLP-IJCNLP)}, pages 4129--4142, Hong Kong,
  China. Association for Computational Linguistics.

\bibitem[{Jin et~al.(2019)Jin, Jin, Zhou, and Szolovits}]{jin2019bert}
Di~Jin, Zhijing Jin, Joey~Tianyi Zhou, and Peter Szolovits. 2019.
\newblock \href {https://arxiv.org/abs/1907.11932} {Is bert really robust? a
  strong baseline for natural language attack on text classification and
  entailment}.
\newblock \emph{arXiv preprint arXiv:1907.11932}.

\bibitem[{Khashabi et~al.(2020)Khashabi, Khot, and
  Sabharwal}]{khashabi-etal-2020-bang}
Daniel Khashabi, Tushar Khot, and Ashish Sabharwal. 2020.
\newblock \href {https://doi.org/10.18653/v1/2020.emnlp-main.12} {More bang for
  your buck: Natural perturbation for robust question answering}.
\newblock In \emph{Proceedings of the 2020 Conference on Empirical Methods in
  Natural Language Processing (EMNLP)}, pages 163--170, Online. Association for
  Computational Linguistics.

\bibitem[{Li et~al.(2016{\natexlab{a}})Li, Chen, Hovy, and
  Jurafsky}]{li-etal-2016-visualizing}
Jiwei Li, Xinlei Chen, Eduard Hovy, and Dan Jurafsky. 2016{\natexlab{a}}.
\newblock \href {https://doi.org/10.18653/v1/N16-1082} {Visualizing and
  understanding neural models in {NLP}}.
\newblock In \emph{Proceedings of the 2016 Conference of the North {A}merican
  Chapter of the Association for Computational Linguistics: Human Language
  Technologies}, pages 681--691, San Diego, California. Association for
  Computational Linguistics.

\bibitem[{Li et~al.(2016{\natexlab{b}})Li, Monroe, and
  Jurafsky}]{word-saliency-li}
Jiwei Li, Will Monroe, and Dan Jurafsky. 2016{\natexlab{b}}.
\newblock \href {https://arxiv.org/abs/1612.08220} {Understanding neural
  networks through representation erasure}.
\newblock \emph{arXiv preprint arXiv:1612.08220}.

\bibitem[{Liu et~al.(2019)Liu, Ott, Goyal, Du, Joshi, Chen, Levy, Lewis,
  Zettlemoyer, and Stoyanov}]{liu2019roberta}
Yinhan Liu, Myle Ott, Naman Goyal, Jingfei Du, Mandar Joshi, Danqi Chen, Omer
  Levy, Mike Lewis, Luke Zettlemoyer, and Veselin Stoyanov. 2019.
\newblock \href {https://arxiv.org/abs/1907.11692} {Roberta: A robustly
  optimized bert pretraining approach}.
\newblock \emph{arXiv preprint arXiv:1907.11692}.

\bibitem[{Maas et~al.(2011)Maas, Daly, Pham, Huang, Ng, and Potts}]{Maas2011}
Andrew~L. Maas, Raymond~E. Daly, Peter~T. Pham, Dan Huang, Andrew~Y. Ng, and
  Christopher Potts. 2011.
\newblock \href {https://www.aclweb.org/anthology/P11-1015} {Learning word
  vectors for sentiment analysis}.
\newblock In \emph{Proceedings of the 49th Annual Meeting of the Association
  for Computational Linguistics: Human Language Technologies}, pages 142--150,
  Portland, Oregon, USA. Association for Computational Linguistics.

\bibitem[{Morris et~al.(2020{\natexlab{a}})Morris, Lifland, Lanchantin, Ji, and
  Qi}]{morris-etal-2020-reevaluating}
John Morris, Eli Lifland, Jack Lanchantin, Yangfeng Ji, and Yanjun Qi.
  2020{\natexlab{a}}.
\newblock \href {https://doi.org/10.18653/v1/2020.findings-emnlp.341}
  {Reevaluating adversarial examples in natural language}.
\newblock In \emph{Findings of the Association for Computational Linguistics:
  EMNLP 2020}, pages 3829--3839, Online. Association for Computational
  Linguistics.

\bibitem[{Morris et~al.(2020{\natexlab{b}})Morris, Lifland, Yoo, Grigsby, Jin,
  and Qi}]{morris-etal-2020-textattack}
John Morris, Eli Lifland, Jin~Yong Yoo, Jake Grigsby, Di~Jin, and Yanjun Qi.
  2020{\natexlab{b}}.
\newblock \href {https://doi.org/10.18653/v1/2020.emnlp-demos.16}
  {{T}ext{A}ttack: A framework for adversarial attacks, data augmentation, and
  adversarial training in {NLP}}.
\newblock In \emph{Proceedings of the 2020 Conference on Empirical Methods in
  Natural Language Processing: System Demonstrations}, pages 119--126, Online.
  Association for Computational Linguistics.

\bibitem[{Mozes et~al.(2021)Mozes, Stenetorp, Kleinberg, and
  Griffin}]{mozes-etal-2021-frequency}
Maximilian Mozes, Pontus Stenetorp, Bennett Kleinberg, and Lewis Griffin. 2021.
\newblock \href {https://aclanthology.org/2021.eacl-main.13} {Frequency-guided
  word substitutions for detecting textual adversarial examples}.
\newblock In \emph{Proceedings of the 16th Conference of the European Chapter
  of the Association for Computational Linguistics: Main Volume}, pages
  171--186, Online. Association for Computational Linguistics.

\bibitem[{Nie et~al.(2019)Nie, Williams, Dinan, Bansal, Weston, and
  Kiela}]{nie2019adversarial}
Yixin Nie, Adina Williams, Emily Dinan, Mohit Bansal, Jason Weston, and Douwe
  Kiela. 2019.
\newblock Adversarial nli: A new benchmark for natural language understanding.
\newblock \emph{arXiv preprint arXiv:1910.14599}.

\bibitem[{Papernot et~al.(2016)Papernot, McDaniel, Swami, and
  Harang}]{papernot2016crafting}
Nicolas Papernot, Patrick McDaniel, Ananthram Swami, and Richard Harang. 2016.
\newblock Crafting adversarial input sequences for recurrent neural networks.
\newblock In \emph{MILCOM 2016-2016 IEEE Military Communications Conference},
  pages 49--54. IEEE.

\bibitem[{Potts et~al.(2020)Potts, Wu, Geiger, and Kiela}]{potts2020dynasent}
Christopher Potts, Zhengxuan Wu, Atticus Geiger, and Douwe Kiela. 2020.
\newblock Dynasent: A dynamic benchmark for sentiment analysis.
\newblock \emph{arXiv preprint arXiv:2012.15349}.

\bibitem[{Ren et~al.(2019)Ren, Deng, He, and Che}]{ren-etal-2019-generating}
Shuhuai Ren, Yihe Deng, Kun He, and Wanxiang Che. 2019.
\newblock \href {https://doi.org/10.18653/v1/P19-1103} {Generating natural
  language adversarial examples through probability weighted word saliency}.
\newblock In \emph{Proceedings of the 57th Annual Meeting of the Association
  for Computational Linguistics}, pages 1085--1097, Florence, Italy.
  Association for Computational Linguistics.

\bibitem[{Ribeiro et~al.(2018)Ribeiro, Singh, and
  Guestrin}]{ribeiro-etal-2018-semantically}
Marco~Tulio Ribeiro, Sameer Singh, and Carlos Guestrin. 2018.
\newblock \href {https://doi.org/10.18653/v1/P18-1079} {Semantically equivalent
  adversarial rules for debugging {NLP} models}.
\newblock In \emph{Proceedings of the 56th Annual Meeting of the Association
  for Computational Linguistics (Volume 1: Long Papers)}, pages 856--865,
  Melbourne, Australia. Association for Computational Linguistics.

\bibitem[{Szegedy et~al.(2013)Szegedy, Zaremba, Sutskever, Bruna, Erhan,
  Goodfellow, and Fergus}]{szegedy2013intriguing}
Christian Szegedy, Wojciech Zaremba, Ilya Sutskever, Joan Bruna, Dumitru Erhan,
  Ian Goodfellow, and Rob Fergus. 2013.
\newblock Intriguing properties of neural networks.
\newblock \emph{arXiv preprint arXiv:1312.6199}.

\bibitem[{Wallace et~al.(2019)Wallace, Rodriguez, Feng, Yamada, and
  Boyd-Graber}]{wallace2019trick}
Eric Wallace, Pedro Rodriguez, Shi Feng, Ikuya Yamada, and Jordan Boyd-Graber.
  2019.
\newblock Trick me if you can: Human-in-the-loop generation of adversarial
  examples for question answering.
\newblock \emph{Transactions of the Association for Computational Linguistics},
  7:387--401.

\bibitem[{Welbl et~al.(2020)Welbl, Minervini, Bartolo, Stenetorp, and
  Riedel}]{welbl-etal-2020-undersensitivity}
Johannes Welbl, Pasquale Minervini, Max Bartolo, Pontus Stenetorp, and
  Sebastian Riedel. 2020.
\newblock \href {https://doi.org/10.18653/v1/2020.findings-emnlp.103}
  {Undersensitivity in neural reading comprehension}.
\newblock In \emph{Findings of the Association for Computational Linguistics:
  EMNLP 2020}, pages 1152--1165, Online. Association for Computational
  Linguistics.

\bibitem[{Wolf et~al.(2019)Wolf, Debut, Sanh, Chaumond, Delangue, Moi, Cistac,
  Rault, Louf, Funtowicz, and Brew}]{Wolf2019HuggingFacesTS}
Thomas Wolf, Lysandre Debut, Victor Sanh, Julien Chaumond, Clement Delangue,
  Anthony Moi, Pierric Cistac, Tim Rault, R'emi Louf, Morgan Funtowicz, and
  Jamie Brew. 2019.
\newblock \href {https://arxiv.org/abs/1910.03771} {Huggingface's transformers:
  State-of-the-art natural language processing}.
\newblock \emph{ArXiv}, abs/1910.03771.

\bibitem[{Xu et~al.(2020)Xu, Zhong, Yepes, and Lau}]{xu2020elephant}
Ying Xu, Xu~Zhong, Antonio Jose~Jimeno Yepes, and Jey~Han Lau. 2020.
\newblock Elephant in the room: An evaluation framework for assessing
  adversarial examples in nlp.
\newblock \emph{arXiv preprint arXiv:2001.07820}.

\bibitem[{Yuan et~al.(2019)Yuan, He, Zhu, and Li}]{yuan2017adversarial}
Xiaoyong Yuan, Pan He, Qile Zhu, and Xiaolin Li. 2019.
\newblock \href {https://doi.org/10.1109/TNNLS.2018.2886017} {Adversarial
  examples: Attacks and defenses for deep learning}.
\newblock \emph{IEEE Transactions on Neural Networks and Learning Systems},
  30(9):2805--2824.

\bibitem[{Zang et~al.(2020)Zang, Qi, Yang, Liu, Zhang, Liu, and
  Sun}]{zang-etal-2020-word}
Yuan Zang, Fanchao Qi, Chenghao Yang, Zhiyuan Liu, Meng Zhang, Qun Liu, and
  Maosong Sun. 2020.
\newblock \href {https://doi.org/10.18653/v1/2020.acl-main.540} {Word-level
  textual adversarial attacking as combinatorial optimization}.
\newblock In \emph{Proceedings of the 58th Annual Meeting of the Association
  for Computational Linguistics}, pages 6066--6080, Online. Association for
  Computational Linguistics.

\bibitem[{Zhou et~al.(2019)Zhou, Jiang, Chang, and
  Wang}]{zhou-etal-2019-learning}
Yichao Zhou, Jyun-Yu Jiang, Kai-Wei Chang, and Wei Wang. 2019.
\newblock \href {https://doi.org/10.18653/v1/D19-1496} {Learning to
  discriminate perturbations for blocking adversarial attacks in text
  classification}.
\newblock In \emph{Proceedings of the 2019 Conference on Empirical Methods in
  Natural Language Processing and the 9th International Joint Conference on
  Natural Language Processing (EMNLP-IJCNLP)}, pages 4904--4913, Hong Kong,
  China. Association for Computational Linguistics.

\end{thebibliography}
\bibliographystyle{acl_natbib}

\appendix

\begin{figure*}[t]
\resizebox{1.0\textwidth}{!}{
    \subfloat[Unperturbed]{{\includegraphics[width=0.32\textwidth]{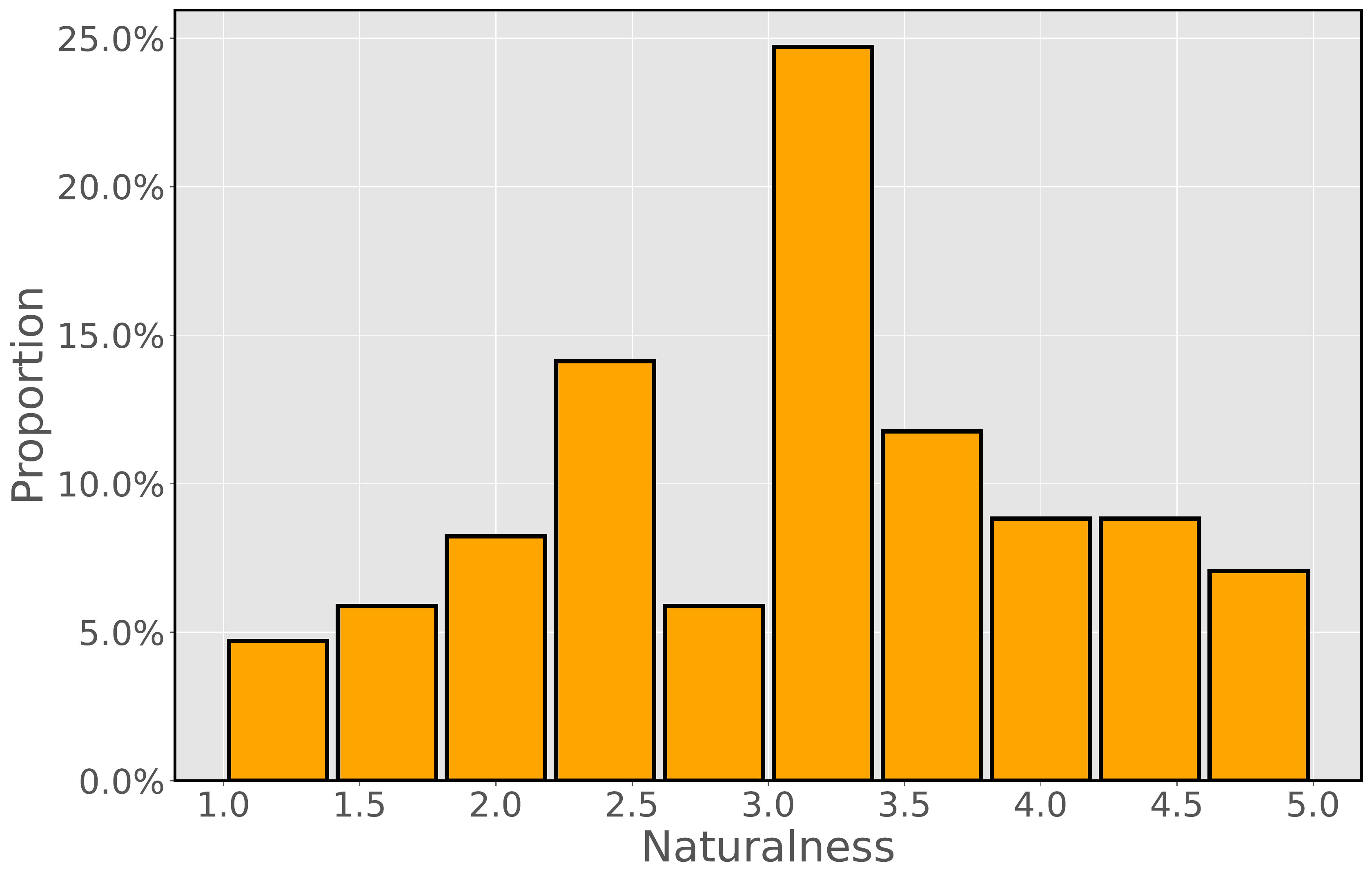}}}
    \subfloat[\textsc{Human}]{{\includegraphics[width=0.32\textwidth]{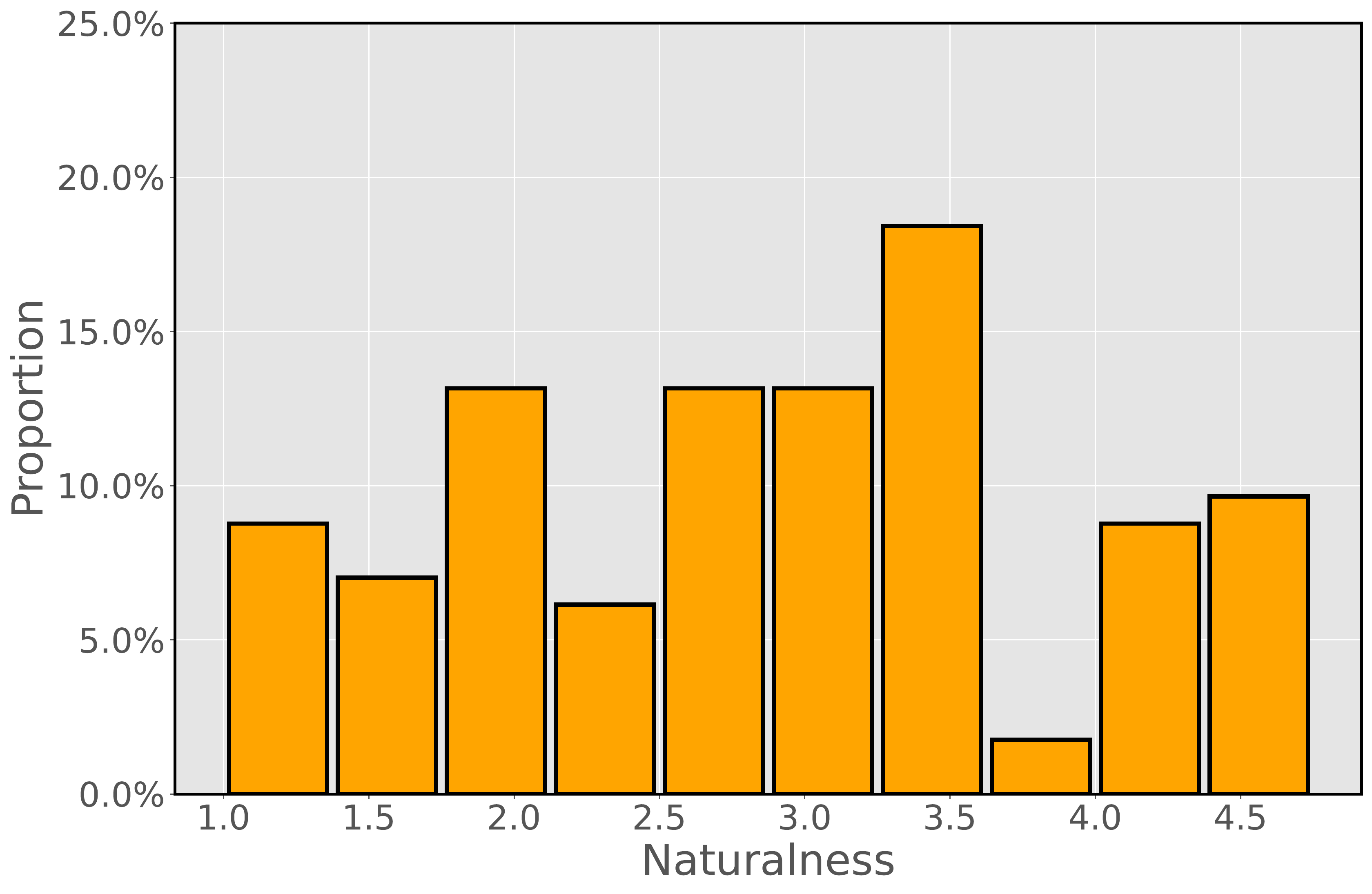}}}
    \subfloat[\textsc{Genetic}]{{\includegraphics[width=0.32\textwidth]{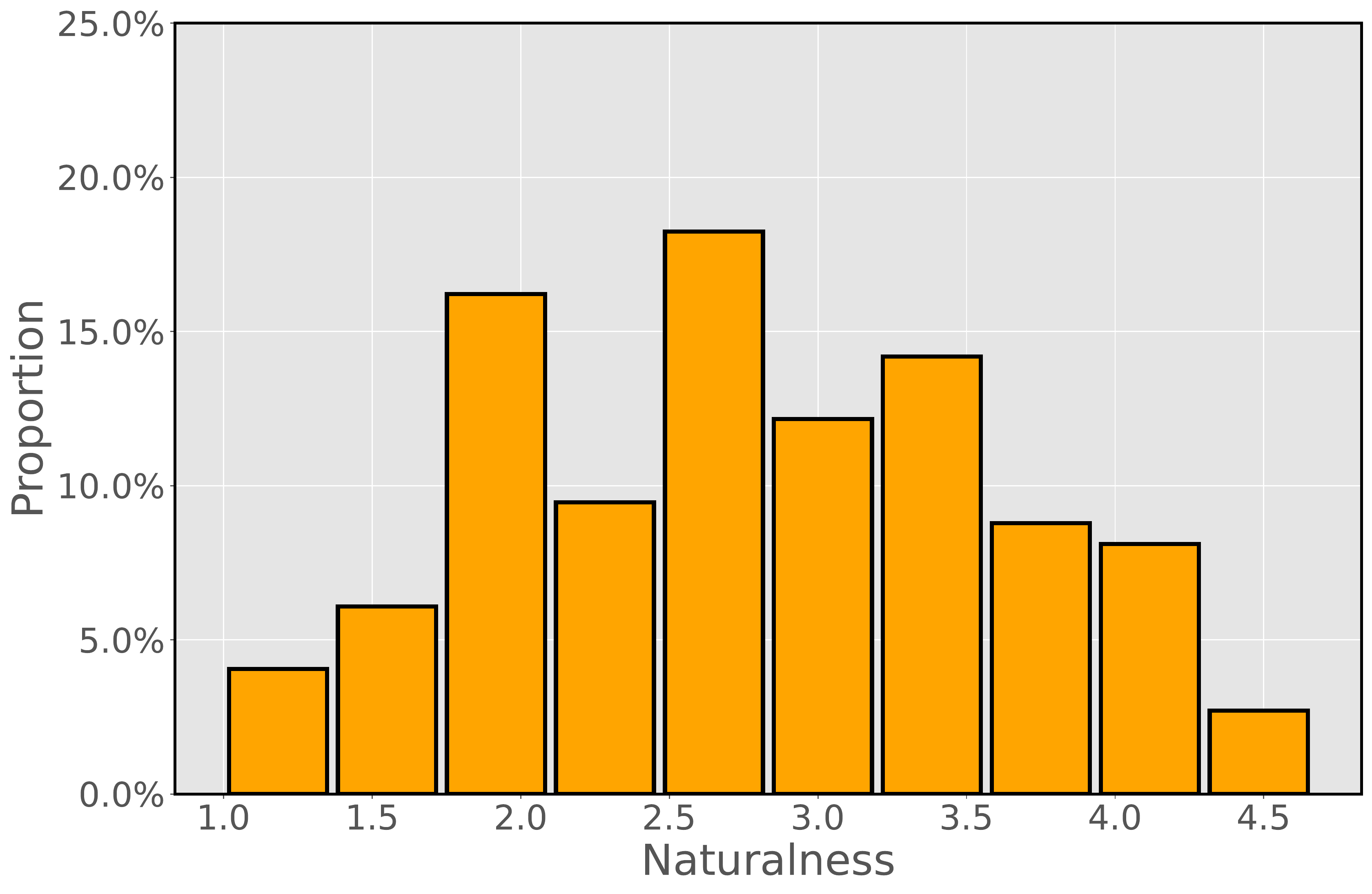}}}
}
\resizebox{1.0\textwidth}{!}{
    \subfloat[\textsc{TextFooler}]{{\includegraphics[width=0.32\textwidth]{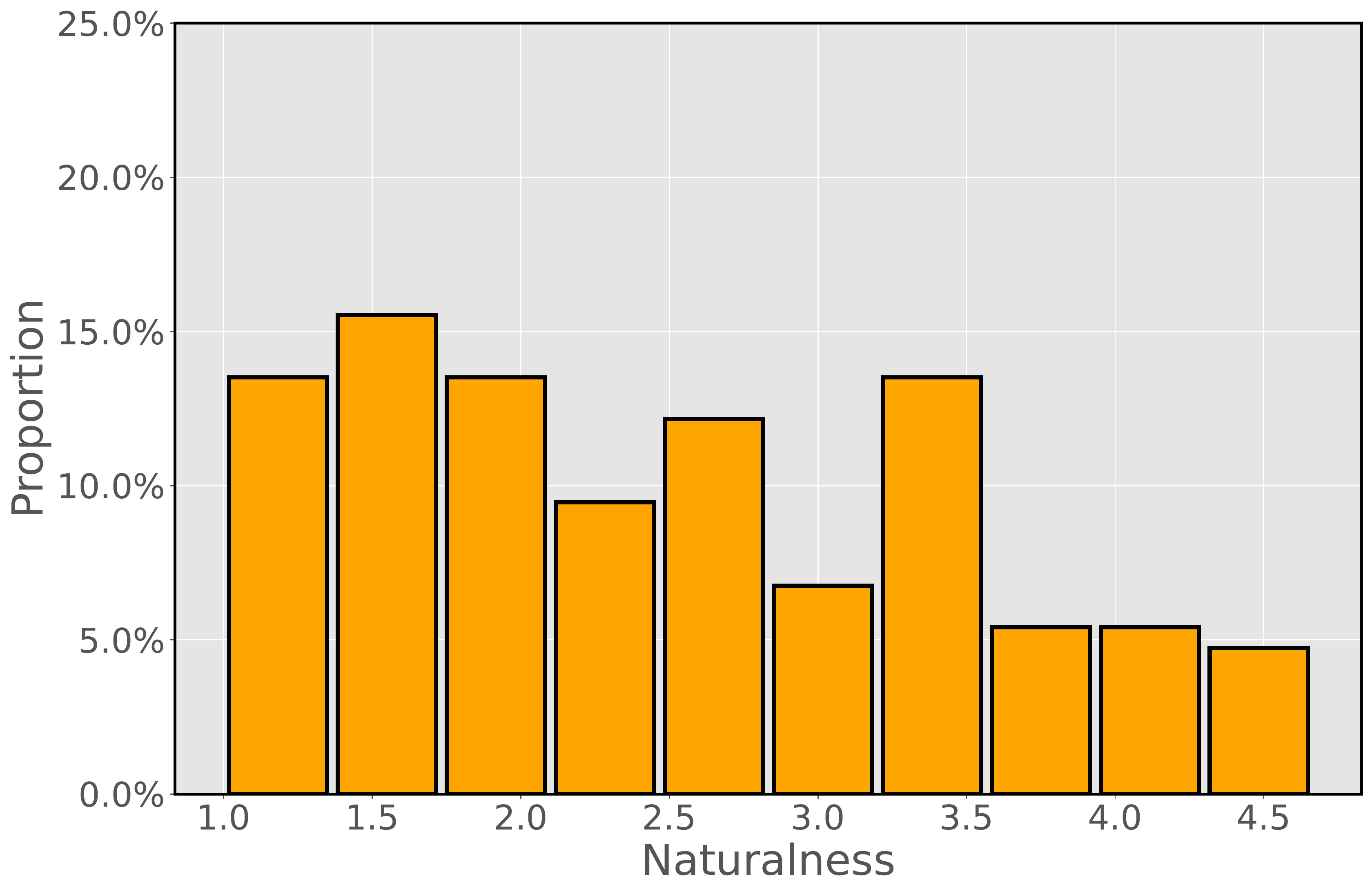}}}
    \subfloat[\textsc{BAE}]{{\includegraphics[width=0.32\textwidth]{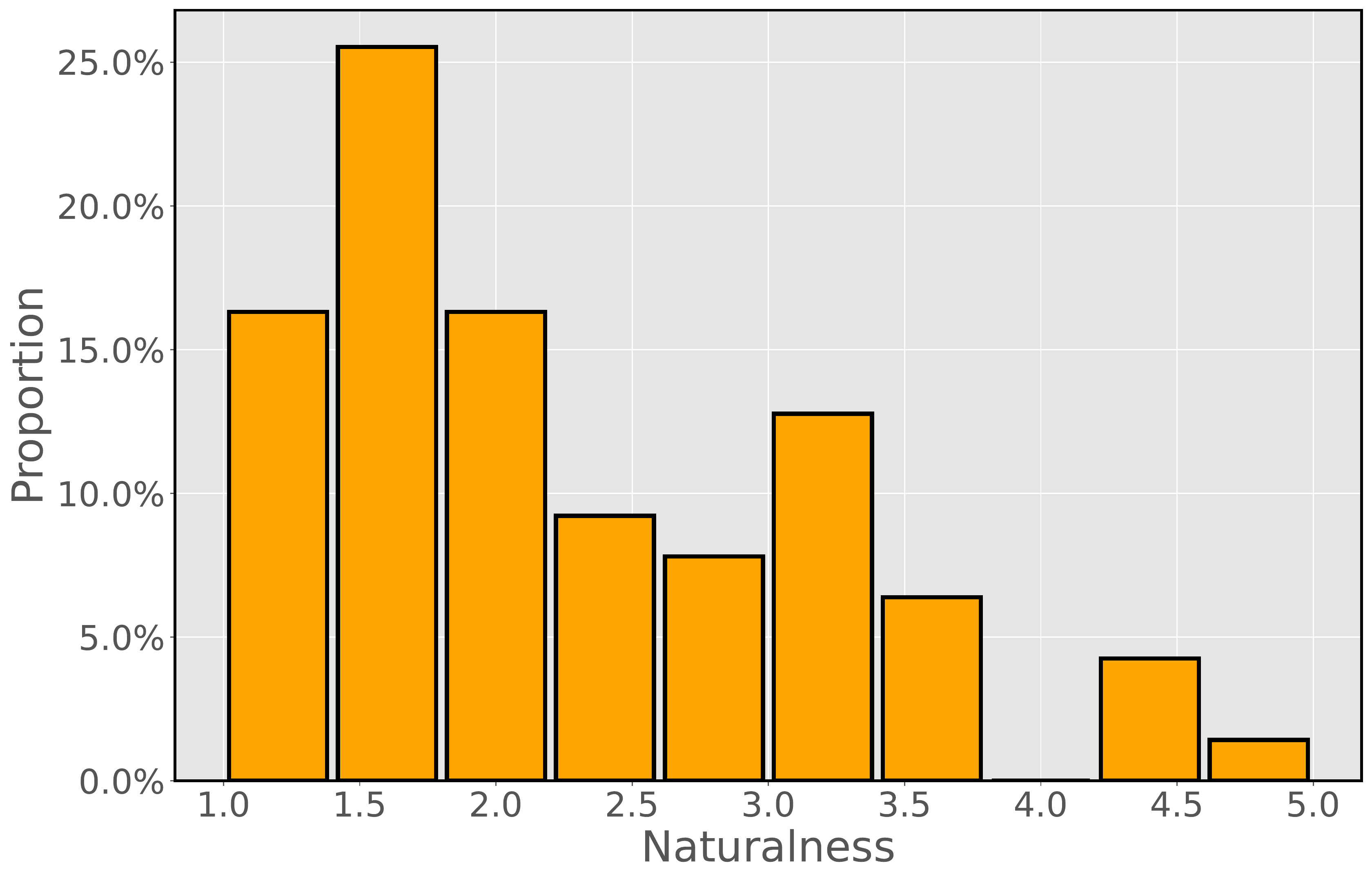}}}
    \subfloat[\textsc{SememePSO}]{{\includegraphics[width=0.32\textwidth]{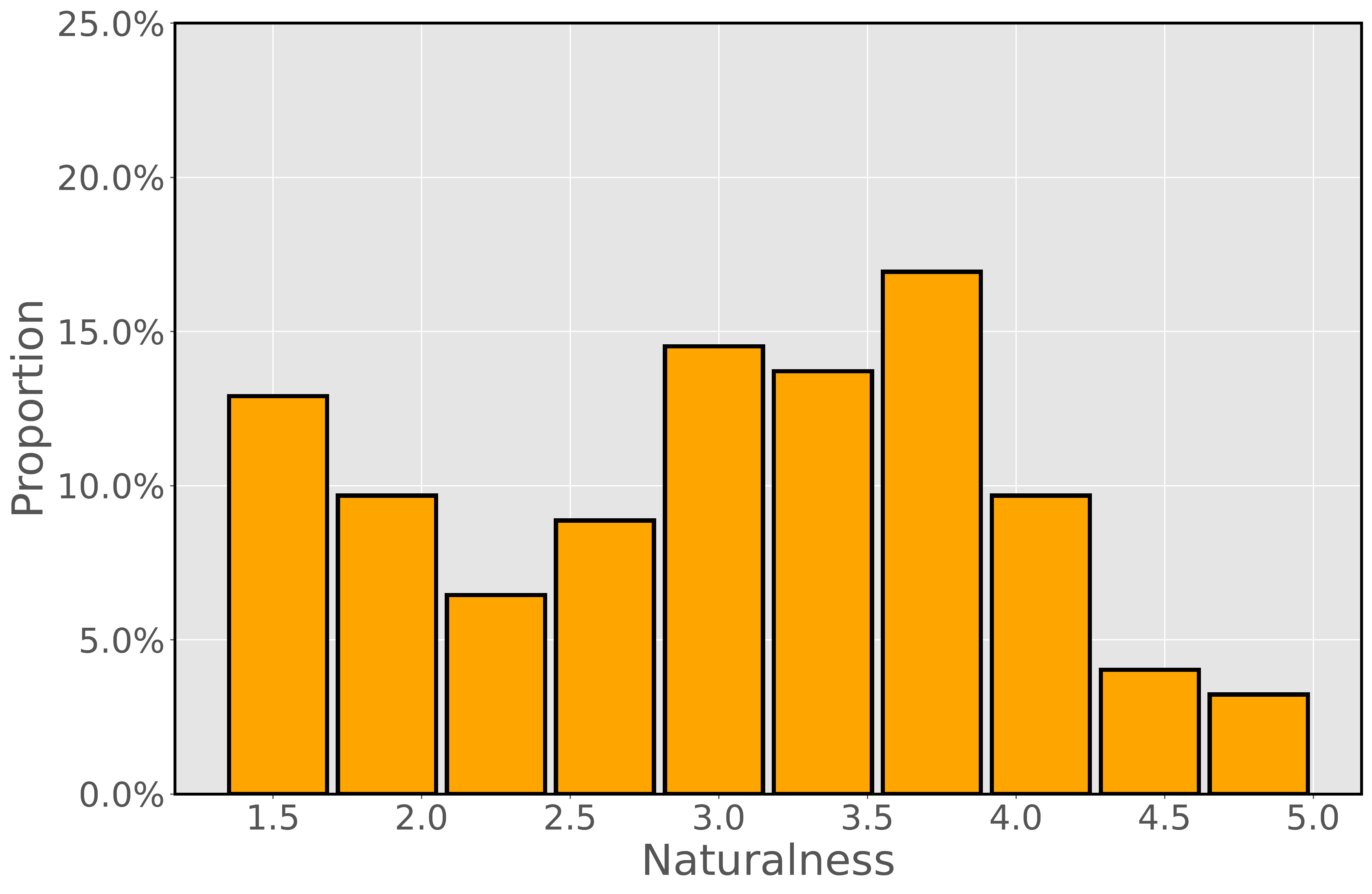}}}
}
\caption{Histograms of the distribution of mean naturalness ratings across examples for each task ($1= \text{very unnatural}$, $5=\text{very natural}$). For all attacks, only the matched adversarial examples (i.e., those that have an agreement between the annotators' and ground truth sentiment label) were considered.}
\label{fig:naturalness-visualization}
\end{figure*}

\begin{table*}[t]
\centering
\footnotesize
\begin{tabular}{{l p{0.49\textwidth} c c c }}
\toprule
\textbf{Attack} & \textbf{Text} & \textbf{Pred.} & \textbf{Naturalness} & \textbf{Sentiment} \\
\midrule
--- & if you are having trouble sleeping or just want to take that nap in the afternoon but just can t seem to drift off, pop in this movie. the only neat thing about this movie are the electric planes. aside from that prepare for some sweet zzzzz s. it boggles the mind how big name stars such as those in this movie can be part of the one of the dullest movies i ve ever seen. now, if you will excuse me, i will finish my nap. & \textit{negative} & 4.5 & 1.9 \\
\textsc{Human} & if you are having \color{red}\textbf{difficulty}\color{black}\, \color{red}\textbf{resting}\color{black}\, or just want to take that \color{red}\textbf{break}\color{black}\, in the afternoon but just can t seem to drift off, pop in this movie. the only \color{red}\textbf{clever}\color{black}\, thing about this movie are the electric planes. aside from that prepare for some \color{red}\textbf{delightful}\color{black}\, zzzzz s. it \color{red}\textbf{amazes}\color{black}\, the mind how big name stars such as those in this movie can be part of the one of the \color{red}\textbf{simplest}\color{black}\, movies i ve ever seen. now, if you will excuse me, i will finish my nap.
 & \textit{positive} & 4.3 & 1.4 \\
\textsc{Genetic} & if you are having trouble \color{red}\textbf{asleep}\color{black}\, or just \color{red}\textbf{wish}\color{black}\, to take that \color{red}\textbf{naps}\color{black}\, in the afternoon but just can t seem to drift off, \color{red}\textbf{dad}\color{black}\, in this movie. the only \color{red}\textbf{groovy}\color{black}\, thing about this \color{red}\textbf{film}\color{black}\, are the \color{red}\textbf{electricity}\color{black}\, \color{red}\textbf{airplanes.}\color{black}\, aside from that prepare for some sweet zzzzz s. it boggles the mind how big \color{red}\textbf{naming}\color{black}\, stars such as those in this movie can be part of the one of the dullest \color{red}\textbf{cinema}\color{black}\, i ve \color{red}\textbf{always}\color{black}\, \color{red}\textbf{observed.}\color{black}\, now, if you will excuse me, i will \color{red}\textbf{complete}\color{black}\, my \color{red}\textbf{naps.}\color{black}\,
 & \textit{negative} & 1.5 & 1.8 \\
\textsc{BAE} & if you are having trouble sleeping or just want to take that nap in the afternoon but just can t seem to drift off, pop in this movie. the only neat thing about this movie are the electric planes. aside from that prepare for some sweet zzzzz s. it boggles the mind how big name stars such as those in this movie can be part of the one of the \color{red}\textbf{liest}\color{black}\, movies i ve ever seen. now, if you will excuse me, i will finish my nap.
 & \textit{positive} & 3.7 & 1.0 \\
\textsc{TextFooler} & if you are having trouble sleeping or just want to take that nap in the afternoon but just can t seem to drift off, pop in this movie. the only neat thing about this movie are the electric planes. aside from that prepare for some sweet zzzzz s. it boggles the mind how big name stars such as those in this movie can be part of the one of the \color{red}\textbf{neatest}\color{black}\, movies i ve ever seen. now, if you will excuse me, i will finish my nap.
 & \textit{positive} & 4.0 & 1.0 \\
\textsc{SememePSO} & if you are having trouble sleeping or just want to take that nap in the afternoon but just can t seem to drift off, pop in this movie. the only neat thing about this movie are the electric planes. aside from that prepare for some sweet zzzzz s. it boggles the mind how big name stars such as those in this movie can be part of the one of the \color{red}\textbf{deepest}\color{black}\, movies i ve ever seen. now, if you will excuse me, i will finish my nap.
 & \textit{positive} & 4.3 & 1.0 \\
\bottomrule
\end{tabular}
\caption{An example movie review from IMDb together with its corresponding adversarial examples.}
\label{fig:illustrations-adversarial-examples-full}
\end{table*}

\end{document}